\documentclass[manuscript, screen, authorversion, nonacm]{acmart}
\AtBeginDocument{%
  \providecommand\BibTeX{{%
    \normalfont B\kern-0.5em{\scshape i\kern-0.25em b}\kern-0.8em\TeX}}}

\acmYear{2024}
\copyrightyear{2024}
\setcopyright{rightsretained}
\acmConference[ACM FAccT '24]{ACM Conference on Fairness, Accountability, and Transparency}{June 3--6, 2024}{Rio de Janeiro, Brazil}
\acmBooktitle{ACM Conference on Fairness, Accountability, and Transparency (ACM FAccT '24), June 3--6, 2024, Rio de Janeiro, Brazil}
\acmDOI{10.1145/3630106.3658975}
\acmISBN{979-8-4007-0450-5/24/06}

\usepackage{makecell} 

\newtheoremstyle{noindent} 
  {3pt} 
  {3pt} 
  {\itshape} 
  {0pt} 
  {\bfseries} 
  {.} 
  {.5em} 
  {} 

\theoremstyle{noindent}
\newtheorem{researchquestion}{Research Question}

\begin{document}

\title[Homogeneity Bias in Large Language Models]{Large Language Models Portray Socially Subordinate Groups as More Homogeneous, Consistent with a Bias Observed in Humans}

\author{Messi H.J. Lee}
\email{hojunlee@wustl.edu}
\orcid{0000-0003-3096-1112}
\affiliation{%
  \institution{Division of Computational and Data Sciences, Washington University in St. Louis}
  \streetaddress{1 Brookings Dr.}
  \city{St. Louis}
  \state{Missouri}
  \country{USA}
  \postcode{63130}
}

\author{Jacob M. Montgomery}
\email{jacob.montgomery@wustl.edu}
\orcid{0000-0001-5632-2437}
\affiliation{%
  \institution{Department of Political Science, Washington University in St. Louis}
  \streetaddress{1 Brookings Dr.}
  \city{St. Louis}
  \state{Missouri}
  \country{USA}
  \postcode{63130}
}

\author{Calvin K. Lai}
\email{calvinlai@wustl.edu}
\orcid{0000-0003-2437-9783}
\affiliation{%
  \institution{Department of Psychological \& Brain Sciences, Washington University in St. Louis}
  \streetaddress{1 Brookings Dr.}
  \city{St. Louis}
  \state{Missouri}
  \country{USA}
  \postcode{63130}
}


\begin{abstract}
Large language models (LLMs) are becoming pervasive in everyday life, yet their propensity to reproduce biases inherited from training data remains a pressing concern. Prior investigations into bias in LLMs have focused on the association of social groups with stereotypical attributes. However, this is only one form of human bias such systems may reproduce. We investigate a new form of bias in LLMs that resembles a social psychological phenomenon where socially subordinate groups are perceived as more homogeneous than socially dominant groups. We had ChatGPT, a state-of-the-art LLM, generate texts about intersectional group identities and compared those texts on measures of homogeneity. We consistently found that ChatGPT portrayed African, Asian, and Hispanic Americans as more homogeneous than White Americans, indicating that the model described racial minority groups with a narrower range of human experience. ChatGPT also portrayed women as more homogeneous than men, but these differences were small. Finally, we found that the effect of gender differed across racial\slash ethnic groups such that the effect of gender was consistent within African and Hispanic Americans but not within Asian and White Americans. We argue that the tendency of LLMs to describe groups as less diverse risks perpetuating stereotypes and discriminatory behavior. 
\end{abstract}







\maketitle

\section{Introduction}
In recent years, the examination of bias in Artificial Intelligence (AI) has garnered significant attention, with multiple studies spotlighting biases in AI systems designed for real-world decision-making \citep[e.g.,][]{dressel_accuracy_2018, ensign_runaway_2017, buolamwini_gender_2018a}. For instance, \citet{buolamwini_gender_2018a} showed that commercial gender classification systems, used in various sectors like marketing, entertainment, security, and healthcare, achieved higher accuracy for lighter-skinned individuals than darker-skinned individuals, and that the disparity was most pronounced within darker-skinned females with error rates high as 34.7\% (as opposed to 0.3\% of lighter-skinned males). This study, along with many others, demonstrated that AI systems, contrary to the expectation that they would be impartial and immune to biases, could show performance disparities for specific groups and reproduce, or even amplify, human biases. 

Natural language processing (NLP) systems are similarly vulnerable to bias. Since the seminal works of \citet{bolukbasi_man_2016} and \citet{caliskan_semantics_2017a} documenting human-like biases within word embedding models, a wide array of studies have found biases within models for coreference resolution \citep{zhao_gender_2018}, text classification \citep{de-arteaga_bias_2019a}, machine translation \citep{prates_assessing_2019, stanovsky_evaluating_2019}, and text generation \citep{lucy_gender_2021, abid_persistent_2021}, among many others. For example, \citet{lucy_gender_2021} showed that GPT-3 would write stories related to family, emotions, and body parts when asked to write about a feminine character whereas it would write stories related to politics, war, sports, and crime when asked to write about a masculine character. Another work by \citet{abid_persistent_2021} showed that GPT-3 would associate Muslims with violence when performing text completions. These studies highlighted the role Large Language Models (LLMs) could play in reproducing and amplifying stereotypical trait associations in their generated content. 

\subsection{Biases beyond trait association}

The above studies not only underscore the potential for LLMs to reproduce and amplify stereotypical trait associations, but they also prompt researchers to question whether LLMs reproduce other human-like biases. One type of bias that remains unexplored in LLMs is perceived homogeneity of groups - the tendency to perceive some social groups as less diverse\slash more homogeneous compared to others. This bias was first studied within the context of intergroup relations where social psychologists found that people tend to perceive members of their outgroup as more homogeneous than members of their ingroup \citep{linville_polarized_1980}. Subsequently, the phenomenon was documented across a wide variety of social distinctions including gender \citep{park_measures_1990}, age \citep{linville_perceived_1989}, race\slash ethnicity \citep{ackerman_they_2006}, college majors \citep{park_role_1992}, and political orientation \citep{quattrone_perception_1980}. However, further exploration revealed that differences in the perceived homogeneity of ingroups and outgroups may instead be attributable to the relative social status and power of groups \citep{guinote_effects_2002, fiske_control_1996, fiske_controlling_1993, lorenzi-cioldi_group_1998, lorenzi-cioldi_they_1993}. These studies found that members of socially dominant groups perceived their out-group(s) as more homogeneous than the ingroup (in line with the typical outgroup homogeneity effect), but that members of socially subordinate groups would perceive their ingroup(s) as more homogeneous than the socially dominant outgroup. Together, these effects suggest that humans have a general tendency to perceive socially subordinate groups as more homogeneous than socially dominant groups. 

Perceived homogeneity (or variability) of groups is a form of stereotyping that has strong implications for prejudice and discrimination. Studies show that viewing a group as more variable reduces other forms of stereotyping \citep{hewstone_perceived_2000a, ryan_effects_1996}, prejudice, and discrimination \citep{brauer_increasing_2011a, er-rafiy_modifying_2013}. As LLMs become increasingly involved in everyday life, it is essential to understand if they perpetuate biases related to perceived homogeneity as they may influence users' perceptions and attitudes towards groups. This investigation is part of a broader discussion on erasure within Natural Language Processing \citep[NLP;][]{dev_measures_2022, dev_harms_2021}, which highlights the lack of adequate representation of social groups in NLP systems. Homogeneous representations of subordinate groups in LLM outputs, or \emph{homogeneity bias}, not only undermine the rich and diverse identities of these groups but also reinforce existing social hierarchies. 

\subsection{Homogeneous narratives of marginalized groups in LLMs}

Recent works in the LLM literature, such as \citet{cheng_marked_2023a} and \citet{cheng_compost_2023}, have highlighted LLMs' tendencies to essentialize and produce positive yet homogeneous narratives of marginalized groups in personas, written descriptions of an individual who identifies with a given social group identity (e.g., ``Imagine you are an Asian woman. Describe yourself.”). \citet{cheng_compost_2023} measure the extent to which these descriptions focus on groups' defining characteristics, often linked to stereotypes, in a manner akin to ``stereotype endorsement," one of three types of measures used to study the outgroup homogeneity effect \citep{ostrom_outgroup_1992}. Building on this, we introduce a new method to assess homogeneity in group representations, akin to ``perceived similarity," which quantifies the degree of similarity in these representations. Furthermore, we extend our analysis to text formats more aligned with everyday use of LLMs (e.g., stories), underscoring the pervasive harm of homogeneity bias. Our findings indicate that homogeneity bias affects not only the content but also the manner in which the narratives are conveyed. 

\subsection{This work}

In this work, we empirically test whether LLMs exhibit bias akin to human perceptions of group homogeneity through an experiment using ChatGPT. We had ChatGPT generate texts about eight different intersectional groups. We looked at four racial\slash ethnic groups - African, Asian, Hispanic, and White Americans - where White Americans were identified as the dominant racial\slash ethnic group \citep{zou_two_2017b}, and we looked at two gender groups - men and women - where men were identified as the dominant gender group \citep{swim_sexism_2009}. If LLMs reproduce this human-like bias, we would expect LLMs to describe members of the socially subordinate group as more homogeneous than those of the socially dominant group. We formalize our pre-registered research questions\footnote{\url{https://osf.io/kxz6b/}} as follows: 

\begin{researchquestion}
Does ChatGPT depict U.S. racial\slash ethnic minority groups (African, Asian, and Hispanic Americans) as more homogeneous compared to the U.S. racial\slash ethnic majority group (White Americans)? 
\end{researchquestion}
\begin{researchquestion}
Does ChatGPT depict the gender minority group (women) as more homogeneous compared to the gender majority group (men)? 
\end{researchquestion}
\begin{researchquestion}
Is the effect of gender on the homogeneity of text generated by ChatGPT consistent across racial\slash ethnic groups?
\end{researchquestion}

\section{Method}

\subsection{Data}

We created a collection of writing prompts asking ChatGPT to write texts about eight intersectional group identities. We included four racial\slash ethnic groups - African, Asian, Hispanic, and White Americans - and two gender groups - men and women. To generate a wide range of comparable content, we considered a variety of text formats such as stories, character descriptions, and biographies. To control for text length, we limited generated text to 30 words.\footnote{ChatGPT did not strictly follow the length requirement. The texts had an average length of 26.61 words (\textit{SD} = 2.70).} The prompts read, ``Write a 30-word [ story about / character description of / biography of / introduction of / social media profile of / synopsis for / narrative of / self-introduction of / tragic story about / funny story about / romantic story about / horror story about / dramatic story about ] a(n) [ African / Asian / Hispanic / White ] American [ man / woman ].” 

We used the OpenAI API, specifically employing the \texttt{gpt-3.5\allowbreak-turbo} model (as of 25 July 2023) to obtain 500 text completions for each prompt. The decision to collect 500 completions stemmed from pilot tests suggesting that a smaller number of completions (i.e., 10 or 100) lead to more instability in our estimates. We used the default parameters of the API,\footnote{\url{https://platform.openai.com/docs/guides/gpt/chat-completions-api}} but made two exceptions: the \texttt{n} parameter, which determines the number of text completions per API request, and the \texttt{role} of the system that determines the model’s behavior (set to ``chatbot'').\footnote{We gathered data in four separate batches, with \texttt{n}s set to 128, 128, 128, and 116 as the API could only process up to 128 generations in each request.} To ensure data quality, we did a keyword-based query to identify and remove 50 out of 52,000 instances where ChatGPT refused to generate the requested texts.\footnote{We provide a breakdown of non-compliant completions by race\slash ethnicity, gender, and text format in Section~\ref{appendix:differential compliance} of the Supplementary Materials. These non-compliant completions were replaced with new ones.}

\subsection{Measure of text homogeneity}

We assessed text homogeneity by calculating the pairwise cosine similarity between sentence embeddings of texts generated for each group. These embeddings are numeric vectors in a multidimensional space that encode the semantic and syntactic information of sentences \citep{conneau_what_2018}. We obtained these embeddings using the second-to-last layer of the BERT-base-uncased model, referred to below as BERT$_{-2}$, following our pre-registered analysis plan. This choice aligned with the default configuration of the \texttt{text} R package \citep[R Version 4.3.1;][]{kjell_textpackage_2023} and reflected the fact that upper layers (i.e. close to last) of the embedding model tend to provide more contextualized representations of language \citep{ethayarajh_how_2019}.

We conducted four sets of additional analyses to evaluate the robustness of our findings to alternative approaches for measuring similarity (these were not pre-registered). We used (1) the third-to-last layer of BERT (BERT$_{-3}$), (2) the second-to-last layer of the larger RoBERTa-base model \citep[][RoBERTa$_{-2}$]{liu_roberta_2019}, (3) the third-to-last layer of RoBERTa (RoBERTa$_{-3}$), and (4) three pre-trained Sentence-BERT models with highest average performance on sentence encoding tasks \citep{reimers_sentencebert_2019}: \texttt{all-mpnet-base-v2}, \texttt{all-distilroberta-v1}, and \texttt{all-MiniLM-L12-v2}.

\begin{table*}[!htbp]
    \caption{Pairs of sentences with the highest and lowest standardized cosine similarity values among stories written about African American men. The cosine similarity values were calculated using BERT$_{-2}$.}
    \centering
    \begin{tabular}
        {p{.4\linewidth}p{.4\linewidth}c}
        \toprule
        \textbf{Sentence 1} & \textbf{Sentence 2} & \textbf{\textit{Std. Cos. Sim.}} \\
        \midrule
        In a world divided by prejudice, he shattered stereotypes with his compassionate heart, empowering others to rise above discrimination and embrace unity. & In a world divided by prejudice, he defied stereotypes with his intelligence and compassion, inspiring others to rise above ignorance and embrace unity. & 1.57 \\ \midrule
        He closed his eyes and took a deep breath, feeling the weight of history on his shoulders. With determination, he stepped forward, ready to redefine his legacy. & An African American man woke up to a world where color no longer mattered, and everyone saw the brilliance in every shade of skin. & $-4.98$ \\
        \bottomrule
    \end{tabular}
    \label{sample sentences}
\end{table*}

After encoding the ChatGPT-generated texts into sentence embeddings, we calculated the cosine similarity between all pairs of the sentence embeddings that were induced for each of the prompts. Cosine similarity is calculated by taking the dot product of two sentence embeddings and dividing it by the product of their magnitudes. The value can range from -1 to 1, where 1 indicates that the two sentences are perfectly identical and where -1 indicates that the two sentences are completely dissimilar. We then standardized this measure for interpretability (subtracting the mean and dividing by the standard deviation). Table~\ref{sample sentences} shows the \emph{most similar} and \emph{least similar} pairs of texts according to the standardized cosine similarity values computed using BERT$_{-2}$. These examples provide some face validity to our measurement strategy as the first sentence pair largely conveys the same message while the second pair does not. To see if this generalizes, we present ten random sentence pairs in Table~\ref{appendix:sample sentences} of the Supplementary Materials. These examples again provide strong face validity for our measurement strategy, with high-scoring pairs appearing to be far more similar than low-scoring pairs. As we generated 500 texts for each prompt, there were 124,750 pairs of sentence embeddings, and hence 124,750 cosine similarity measurements corresponding to each prompt.

\subsection{Testing group differences}

Following the pre-registered analysis plan, we used linear mixed-effects models with functions from the \texttt{lme4} \citep{bates_fitting_2015} and \texttt{lmerTest} \citep{kuznetsova_lmertest_2017} R packages. In the models, we included race\slash ethnicity, gender, and their interactions as fixed effects and text format as random intercepts. Text format was included as random intercepts instead of random slopes because we expected the cosine similarity baseline to vary across text formats,\footnote{Text formats like self-introduction, for example, may be more similar to each other than other text formats given that self-introductions are likely to share a common structure or content that constitutes an introduction.} but we did not expect the magnitude and direction of race\slash ethnicity and gender to vary across text format.\footnote{When fitting linear mixed-effects models, we turned off derivative calculations that could slow down the model fitting process and used the \texttt{nmkbw} optimizer made available by the \texttt{lme4} R package.} 

We also fitted additional un-pre-registered models to facilitate interpretation of race\slash ethnicity and gender fixed effects in the presence of interactions \citep{brown_introduction_2021}. We fitted mixed-effects models where (1) race\slash ethnicity was the only fixed effect (``Race\slash Ethnicity model"), (2) gender was the only fixed effect (``Gender model"), and (3) race\slash ethnicity and gender were both fixed effects (``Race\slash Ethnicity \& Gender model"). These models allowed for easier interpretation and led to the same substantive conclusions. Subsequently, we used the pre-registered mixed-effects model (``Interaction model") to interpret the interaction effect. 

We used the \texttt{afex} R package \citep{singmann_afex_2024} to conduct likelihood-ratio tests to determine if the models including the fixed effects of race\slash ethnicity, gender, and their interactions provided better fits for the data than those without. To determine the magnitude and direction of race\slash ethnicity and gender, we examined the summary outputs of the Race\slash Ethnicity and Gender models. Finally, to examine the interaction effects, we used the \texttt{emmeans} R package \citep{lenth_emmeans_2024} to conduct pairwise comparisons of estimated marginal means between gender groups within the same racial\slash ethnic groups. In all models, White Americans and men served as reference categories.\footnote{Code is available at \url{https://github.com/lee-messi/Homogeneity-Bias-in-LLMs}}

\section{Results}

In Table~\ref{descriptive statistics}, we present the means and standard deviations of the standardized cosine similarity values for the eight intersectional groups, computed using BERT$_{-2}$. 

\begin{table}[!htbp]
    \centering
    \caption{Descriptive statistics of the standardized cosine similarity values for the eight intersectional groups. Cosine similarity computations were performed using BERT$_{-2}$ and were then standardized for better interpretability.}
    \begin{tabular}{lcccc}
        \toprule
        \textbf{Race\slash Ethnicity} & \textbf{Gender} & \textbf{\textit{N}} & \textbf{\textit{Mean}} & \textbf{\textit{St. Dev.}} \\ \midrule
        African Americans & Men & 124,750 & 0.12 & 0.79 \\
        & Women & 124,750 & 0.13 & 0.86 \\ \midrule
        Asian Americans & Men & 124,750 & 0.10 & 0.83 \\
        & Women & 124,750 & 0.11 & 0.87 \\ \midrule
        Hispanic Americans & Men & 124,750 & -0.09 & 1.34 \\
        & Women & 124,750 & 0.04 & 1.25 \\ \midrule
        White Americans & Men & 124,750 & -0.21 & 0.89 \\
        & Women & 124,750 & -0.21 & 0.95 \\
        \bottomrule
    \end{tabular}
    \label{descriptive statistics}
\end{table}

\subsection{Main effect of race\slash ethnicity}

\begin{figure}[!htbp]
  \centering
  \includegraphics[width = \linewidth]{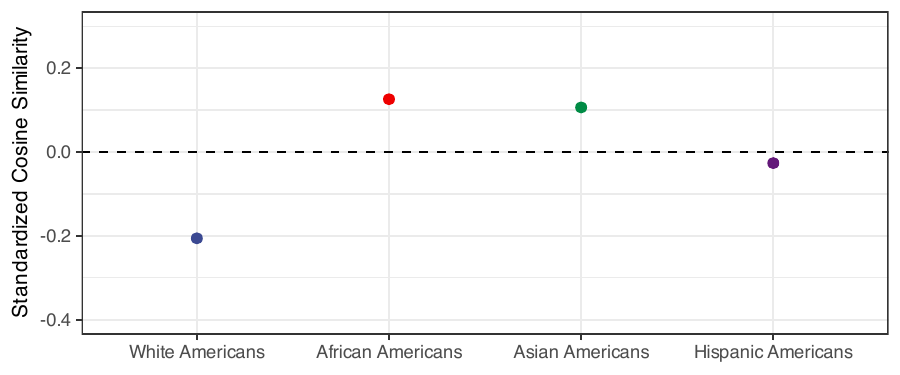}
  \caption{Mean standardized cosine similarity values of the four racial\slash ethnic groups using BERT$_{-2}$. Error bars were omitted as confidence intervals were all smaller than 0.001.}
  \Description{Mean standardized cosine similarity values of the four racial\slash ethnic groups using BERT$_{-2}$. Error bars were omitted as confidence intervals were all smaller than 0.001.}
  \label{race-effect}
\end{figure}

ChatGPT-generated texts about the subordinate racial\slash ethnic groups were more homogeneous than those about the dominant racial\slash ethnic group (see Figure~\ref{race-effect}). The Race\slash Ethnicity model (Column 1 in Table~\ref{BERT-2 output}) showed that the standardized cosine similarity values of African, Asian, and Hispanic Americans were each 0.33 (\textit{SE} < .001, \textit{t}(12,973,984) = 508.81), 0.31 (\textit{SE} < .001, \textit{t}(12,973,984) = 478.74), and 0.18 (\textit{SE} < .001, \textit{t}(12,973,984) = 275.05) standard deviations greater than those of White Americans. In addition, the likelihood-ratio test showed that the model including race\slash ethnicity provided a better fit for the data than that without it, as indicated by the chi-squared statistics for the analysis using BERT$_{-2}$ ($\chi^2$(3) = 326701.07, \textit{p} $<$ .001; see Table~\ref{appendix:lrt-output}). These findings replicated across all six alternative measurement strategies. For results of the likelihood ratio tests, see Table~\ref{appendix:lrt-output}, and for summary outputs of the mixed effects models, see Tables~\ref{appendix:BERT-3 output}-\ref{appendix:all-MiniLM-L12-v2 output}. 

\begin{table*}
    \caption{Summary output of mixed effects models using cosine similarity values from BERT$_{-2}$. Positive coefficients indicate greater pairwise cosine similarity and thus more homogeneity compared to the baseline categories - White Americans and men.}
    \label{BERT-2 output}
    \centering
        \begin{tabular}{lccccccc}
        \toprule
        \multicolumn{1}{c}{}& 
        \multicolumn{7}{c}{BERT$_{-2}$}\\
        \cmidrule{2-8} 
        & \makecell{Race\slash Ethnicity\\model} & & \makecell{Gender\\model} & & \makecell{Race\slash Ethnicity,\\Gender\\model} & & \makecell{Interaction\\model} \\
        \cmidrule{2-2} \cmidrule{4-4} \cmidrule{6-6} \cmidrule{8-8} 
        \textbf{Fixed Effects} & & & & & & & \\ [1ex]
        Intercept & $-0.21$ & & $-0.018$ & & $-0.22$ & & $-0.21$ \\ 
         & (0.16) & & (0.16) & & (0.16) & & (0.16) \\ [1ex]
        African Americans & 0.33$^{*}$ & & & & 0.33$^{*}$ & & 0.33$^{*}$ \\ 
         & (0.00065) & & & & (0.00065) & & (0.00092) \\ [1ex]
        Asian Americans & 0.31$^{*}$ & & & & 0.31$^{*}$ & & 0.31$^{*}$ \\
         & (0.00065) & & & & (0.00065) & & (0.00092) \\ [1ex]
        Hispanic Americans & 0.18$^{*}$ & & & & 0.18$^{*}$ & & 0.12$^{*}$ \\ 
         & (0.00065) & & & & (0.00065) & & (0.00092) \\ [1ex]
        Women & & & 0.037$^{*}$ & & 0.037$^{*}$ & & 0.00021\\ 
         & & & (0.00047) & & (0.00046) & & (0.00092)\\ [1ex]
        African Americans $\times$ Women & & & & & & & 0.0097$^{*}$ \\
         & & & & & & & (0.0013) \\ [1ex]
        Asian Americans $\times$ Women & & & & & & & 0.013$^{*}$ \\
         & & & & & & & (0.0013) \\ [1ex]
        Hispanic Americans $\times$ Women & & & & & & & 0.12$^{*}$ \\
         & & & & & & & (0.0013) \\ [1ex]
        \textbf{Random Effects} ($\mathbf{\sigma^2}$) & & & & \\ [1ex]
        Text Format Intercept & 0.32 & & 0.32 & & 0.32 & & 0.32 \\ [1ex]
        Residual & 0.69 & & 0.71 & & 0.69 & & 0.69 \\ [1ex]
        \midrule
        Observations & 12,974,000 & & 12,974,000 & & 12,974,000 & & 12,974,000  \\
        Log likelihood & $-15,985,323$ & & $-16,145,340$ & & $-15,982,157$ & & $-15,976,230$ \\ 
        \bottomrule
        \addlinespace[0.5ex]
        \multicolumn{1}{l}{\footnotesize{*$p < .001$}}
        \end{tabular}
\end{table*}

\subsection{Main effect of gender}

\begin{figure}[!htbp]
  \centering
  \includegraphics[width = \linewidth]{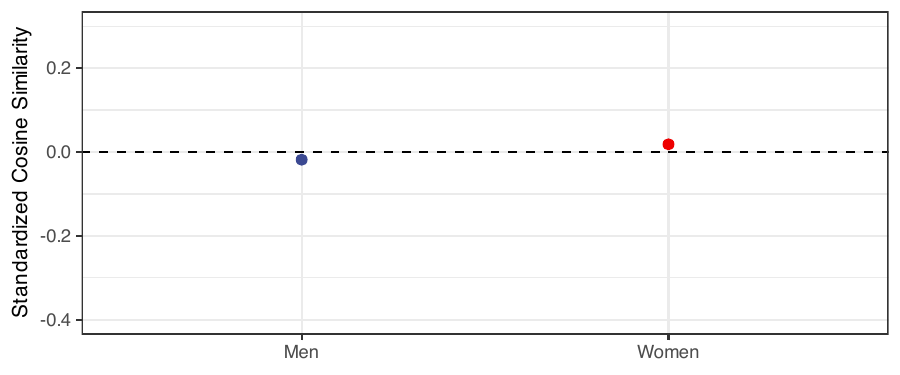}
  \caption{Standardized cosine similarity values of the two gender groups using BERT$_{-2}$. Error bars were omitted as confidence intervals were all smaller than 0.001.}
  \Description{Standardized cosine similarity values of the two gender groups using BERT$_{-2}$. Error bars were omitted as confidence intervals were all smaller than 0.001.}
  \label{gender-effect}
\end{figure}

ChatGPT-generated texts about the subordinate gender group (i.e., women) were also more homogeneous than those about the dominant gender group (men), although the differences were modest (see Figure~\ref{gender-effect}). The Gender model in Table~\ref{BERT-2 output} showed that the cosine similarity values of women were 0.037 (\textit{SE} $<$ .001, \textit{t}(12,973,986) = 78.68) standard deviations greater than those of men.\footnote{The base term for gender in the Interaction model (Column 4 of Table~\ref{BERT-2 output}) was not significant, but this does not mean that gender had no effect. Rather, this indicates that gender had no measurable effect within White Americans (the reference category). We discuss this further in the next section.} Furthermore, the likelihood-ratio test found that the model including the gender term provided a better fit for the data than that without it, as indicated by the chi-squared statistics for the analysis using BERT$_{-2}$ ($\chi^2$(1) = 6352.47, \textit{p} $<$ .001; see Table~\ref{appendix:lrt-output}). These findings replicated across all six alternative measurement strategies. For results of the likelihood ratio tests, see Table~\ref{appendix:lrt-output}, and for summary outputs of the mixed effects models, see Tables~\ref{appendix:BERT-3 output}-\ref{appendix:all-MiniLM-L12-v2 output}. However, we note that, although statistically significant, these results indicated that the impact of gender was substantially smaller than that of race\slash ethnicity.

\subsection{Interaction effect}

\begin{figure}[ht]
  \centering
  \includegraphics[width = \linewidth]{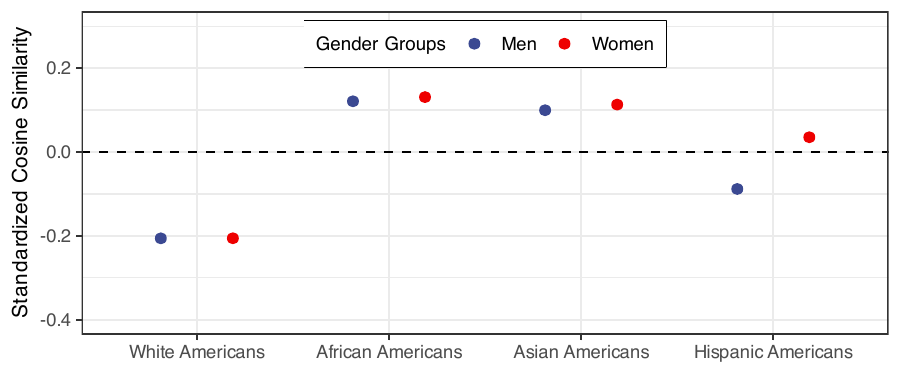}
  \caption{Standardized cosine similarity values of all eight intersectional groups using BERT$_{-2}$. Error bars were omitted as confidence intervals were all smaller than 0.001.}
  \Description{Standardized cosine similarity values of all eight intersectional groups using BERT$_{-2}$. Error bars were omitted as confidence intervals were all smaller than 0.001.}
  \label{interaction-effect}
\end{figure}

The effect of gender on the homogeneity of ChatGPT-generated text differed between racial\slash ethnic groups. Pairwise comparisons of estimated marginal means revealed that African, Asian, and Hispanic American women each held greater cosine similarity values than their male counterparts (\textit{z}s = 10.79, 14.54, 133.86, \textit{p}s $<$ .001), but there was no significant difference between White American men and women (\textit{z} = 0.23, \textit{p} = .82; see Table~\ref{appendix:pairwise-comparison-output} and Figure~\ref{interaction-effect}). The likelihood-ratio test found that the model including the interaction term provided a better fit for the data than that without it, as indicated by the chi-squared statistics for the analysis using BERT$_{-2}$ ($\chi^2$(3) = 11888.15, \textit{p} $<$ .001; see Table~\ref{appendix:lrt-output}). 

\begin{figure}[!htbp]
  \centering
  \includegraphics[width = \linewidth]{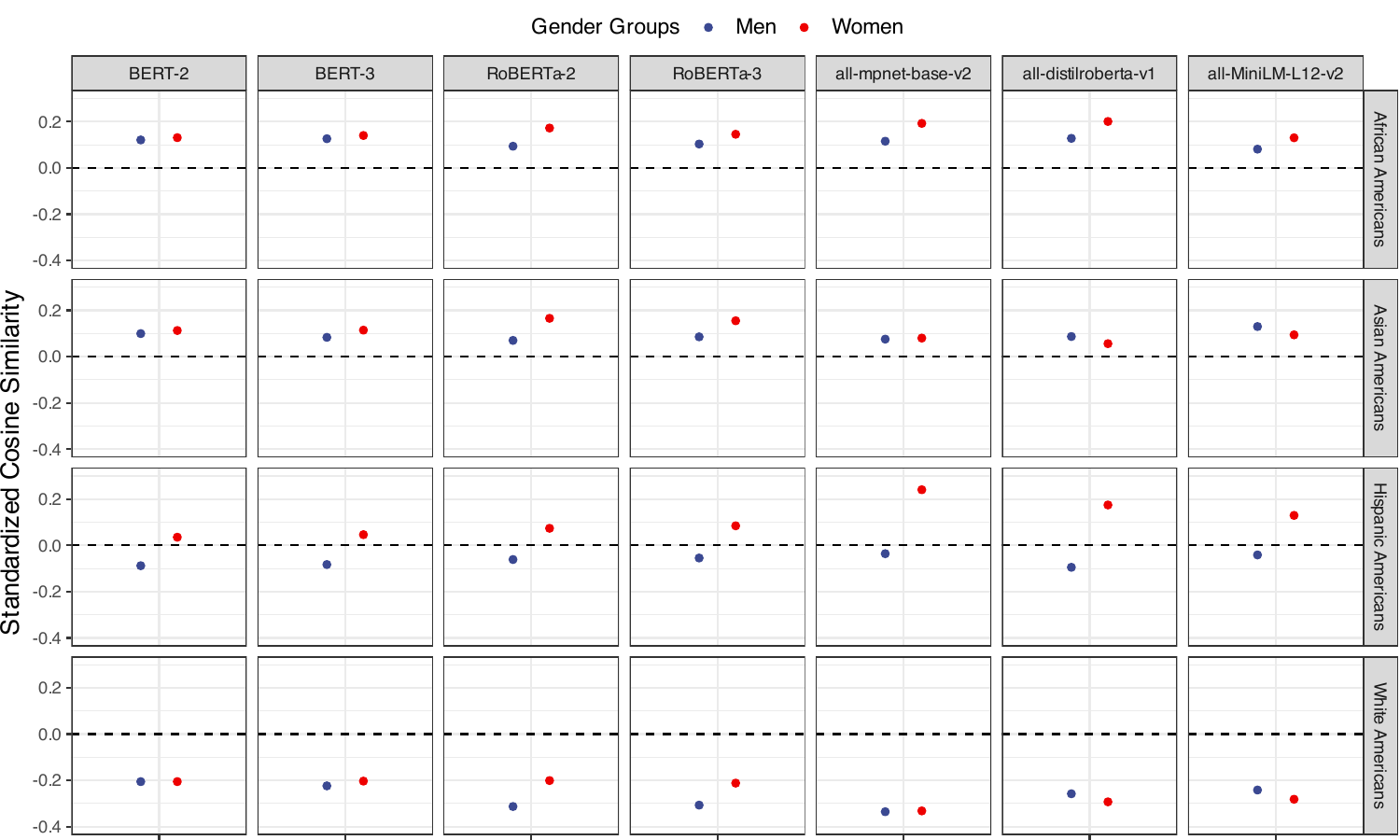}
  \caption{Standardized cosine similarity values of all eight intersectional groups using all seven model specifications. Error bars were omitted as confidence intervals were all smaller than 0.001.}
  \Description{Standardized cosine similarity values of all eight intersectional groups using all seven model specifications. Error bars were omitted as confidence intervals were all smaller than 0.001.}
  \label{all-interactions}
\end{figure}

We observed slight variations in the effects of gender within individual racial\slash ethnic groups when alternative measurement strategies involving BERT and RoBERTa were used (see Figure~\ref{all-interactions}). Examining the results in Table~\ref{appendix:pairwise-comparison-output}, African American women held greater cosine similarity values than their male counterpart (\textit{z}s = 15.34, 82.55, 44.27, \textit{p}s $<$ .001), Asian American women held greater cosine similarity values than their male counterpart (\textit{z}s = 34.32, 100.39, 72.79, \textit{p}s $<$ .001), and Hispanic American women held greater cosine similarity values than their male counterpart (\textit{z}s = 142.07, 141.82, 145.79, \textit{p}s $<$ .001). However, unlike the pre-registered analysis reported in Table~\ref{appendix:pairwise-comparison-output}, White American women also held greater cosine similarity values than their male counterpart (\textit{z}s = 22.61, 117.75, 99.70, \textit{p}s $<$ .001).\footnote{The likelihood-ratio tests shown in Table~\ref{appendix:lrt-output} also indicate the models including the interaction term provided better fits for the data than those without it, as indicated by the chi-squared statistics for the analysis using BERT$_{-3}$ ($\chi^2$(3) = 10618.63, \textit{p} $<$ .001), RoBERTa-2 ($\chi^2$(3) = 1917.00, \textit{p} $<$ .001), and RoBERTa-3 ($\chi^2$(3) = 5591.13, \textit{p} $<$ .001).}

We observed more variations in the effects of gender within individual racial\slash ethnic groups when alternative measurement strategies involving Sentence-BERT were used. Consistent with the pre-registered analysis, African American women held greater cosine similarity values than their male counterpart (\textit{z}s = 98.34, 95.25, 64.65, \textit{p}s $<$ .001), and Hispanic American women held greater cosine similarity values than their male counterpart (\textit{z}s = 352.72, 351.10, 224.90, \textit{p}s $<$ .001). However, the direction of the effect of gender within Asian Americans differed across models (\textit{z}s = 5.81, $-40.29$, $-47.15$, \textit{p}s $<$ .001). Similarly, the direction of the effect of gender within White Americans differed across models (\textit{z}s = 4.61, $-45.44$, $-52.52$, \textit{p}s $<$ .001). All in all, the effect of gender was consistent in one direction within African and Hispanic Americans but not within Asian and White Americans.\footnote{Again, the likelihood-ratio tests found that the models including the interaction term provided better fits for the data than those without it, as indicated by the chi-squared statistics for all-mpnet-base-v2 ($\chi^2$(3) = 80643.97, \textit{p} $<$ .001), all-distilroberta-v1 ($\chi^2$(3) = 103107.16, \textit{p} $<$ .001), and all-MiniLM-L12-v2 ($\chi^2$(3) = 50627.14, \textit{p} $<$ .001).} 

\subsection{Homogeneity bias and topical alignment}

In Section~\ref{appendix:topical alignment} of the Supplementary Materials, we conducted two un-pre-registered follow-up studies and an exploratory analysis to unpack the source of homogeneity bias as measured from cosine similarity of sentence embeddings. We explored whether topical alignment, defined as the frequency of shared topics in texts about specific groups, might account for the observed homogeneity bias. We found that the subordinate racial\slash ethnic groups were discussed more often in terms of hardship and adversity, but we also found that subordinate racial\slash ethnic groups were portrayed as more homogeneous than the dominant racial\slash ethnic group in texts that (1) \emph{were not} about hardship and adversity, and (2) \emph{were} about hardship and adversity. These results indicated that the observed homogeneity bias was partly attributable to shared topics, but that this bias could not be fully explained by topical alignment alone as homogeneity bias also existed \emph{within} topics. This suggested that the bias may also be attributed to other elements, such as alignment of semantic meaning or syntax, aspects that sentence embeddings capture but topic models do not.

\section{Discussion}

We found that both race\slash ethnicity and gender influence the homogeneity of group representations in LLM-generated text. We consistently found that ChatGPT portrayed socially subordinate racial\slash ethnic groups (African, Asian, and Hispanic Americans) as more homogeneous than the socially dominant racial\slash ethnic group (White Americans). We consistently found that ChatGPT portrayed the socially subordinate gender group (women) as more homogeneous than the socially dominant gender group (men) and that the effect of gender was smaller than that of race\slash ethnicity. Finally, we found that the effect of gender differed across racial\slash ethnic groups such that the effect of gender was consistent within African and Hispanic Americans but not within Asian and White Americans. These results underscore the interplay between race\slash ethnicity and gender, emphasizing the importance of considering intersectionality when investigating representational biases in large language models.

\subsection{Where might these biases be coming from?}

LLMs reproduce biases embedded in their training data. As such, it is likely that homogeneous representations of subordinate groups in texts generated by LLMs are also reproductions of bias in the training data. Given the size and opacity of LLM training data \citep{bender_dangers_2021}, it is difficult to confirm the presence of homogeneity bias within LLM training data. Therefore, we speculate on potential sources of homogeneity bias in the training data. 

One potential source is selection bias where certain groups are over-represented in LLM training data \citep{shah_predictive_2020}. As Tripodi’s study of Wikipedia text \citep{tripodi_ms_2023} would suggest, some groups are more frequently discussed in the training data of LLMs. Higher frequency of a group in the training data would result in the LLM generating more diverse text for that group as it allows the model to access a broader and varied set of examples to learn from. Future work should explore how different levels of group representation in training data affect homogeneity of LLM-generated text, perhaps by examining the bias in two otherwise equivalent LLMs, one that is trained on a gender- or race-balanced corpus, for example, and another that is not. Establishing this causal link would guide efforts to mitigate this bias in LLMs, ensuring fair and diverse representations of groups.

Another potential source is stereotypical trait associations in training data \citep{shah_predictive_2020}. Training data of LLMs reflect the dominant group's worldview \citep{bender_dangers_2021}, which, as \citet{fiske_controlling_1993} suggests, is more prone to stereotyping socially subordinate groups according to certain traits. This tendency in LLM training data can lead to subordinate groups being described according to a stereotypical trait, reducing the diversity of words and ideas that LLMs associate with these groups. Future work should explore how stereotypical trait associations in training data affects homogeneity of group representations in LLM-generated text, providing insights into the underlying dynamics of LLM training and aiding the development of fairer and less biased language models.

\section{Limitations and future directions}

We documented the bias using 30-word texts generated by ChatGPT because they serve as a good unit of text for an initial exploration and facilitates the measurement of text similarity using sentence embeddings. However, ChatGPT-generated responses are rarely 30-words long. Consequently, this work would benefit from future work exploring the bias in longer forms of text. Considering the coherence and interconnectedness of longer forms of text, we expect the bias to amplify across sentences and paragraphs and manifest similarly, if not more prominently, in extended texts. By extending our investigation to longer and diverse forms of text, we could strengthen the overall understanding of the observed bias and its implications beyond the confines of 30-word texts.

Second, we used group labels to indicate group identities. However, identities can be signaled in many different ways, such as through names (e.g., Jane Lopez) and other labels (e.g., Mexican Americans). LLM performance is heavily influenced by the prompts used \citep{zhao_calibrate_2021}, so future work should explore the generalizability of these findings using alternative identity signaling methods. These explorations could potentially tackle the ``(un)markedness" issue \citep[see][]{blodgett_stereotyping_2021} in our prompt design where prompts using ``White American” and ``man" may be deemed unsuited for comparison given that these identities tend to be unmarked in discourse \citep{bucholtz_language_2005}. Nevertheless, the fact that these typically unmarked terms yielded more varied representations suggests that we might be underestimating the extent of homogeneity bias in LLMs and that actual homogeneity bias could be even more significant.\footnote{If terms like ``White American" and ``man" are typically unmarked, by explicitly signaling their group identity, we are capturing a narrower representation of the groups where their group identities are explicitly mentioned in the training data. Hence, we would expect representations from these prompts to be more homogeneous than their actual representations.} 

Third, we acknowledge the limited scope of group identities explored in our study. We prioritized groups that reflected some of the largest subsets of the U.S. population. Including smaller groups, such as Native or Middle Eastern Americans, or people with non-binary gender identities, would have expanded the generalizability of our findings. Given that homogeneity bias may stem from under-representation in the training data, we speculate smaller groups may show even stronger evidence of homogeneity bias than some of the groups we examined in the current study.

\section{Conclusion}

We uncovered a new type of bias in Large Language Models (LLMs) that pertains to the variability in representations of socially subordinate and dominant groups. Our findings indicated that LLMs depict socially subordinate groups as more homogeneous than the dominant group, although the effect of gender was smaller than the effect of race\slash ethnicity. Moreover, the interaction between race\slash ethnicity and gender influenced this bias, with the effect of gender being consistent within African and Hispanic Americans but not within Asian and White Americans. The presence of this bias in LLMs raises concerns about the potential erasure of diverse experiences among subordinate groups and the reinforcement of stereotypes. Future research should explore strategies to mitigate this bias in LLMs, aiming to enhance fairness, equity, and inclusivity in their generated content.

\bibliographystyle{ACM-Reference-Format}
\bibliography{homogeneity}

\appendix

\setcounter{table}{0}
\setcounter{figure}{0}

\renewcommand{\thetable}{A\arabic{table}}
\renewcommand{\thefigure}{A\arabic{figure}}

\clearpage

\section{Supplementary Materials}
\subsection{Face validity of the cosine similarity measurements}
\label{appendix:face validity}

To demonstrate the face validity of the cosine similarity measurements, we provide ten randomly selected pairs from ChatGPT-generated stories about a White American man, arranged in descending order of cosine similarity in Table \ref{appendix:sample sentences}. As one progresses through the table, it becomes evident that the overlap in semantic meaning diminishes with the decreasing cosine similarity values. 

\begin{table*}[!htbp]
    \caption{Ten randomly selected pairs of stories about a White American man arranged in descending order of cosine similarity. To better distinguish the cosine similarity values, we report cosine similarity values up to four decimal places.} 
    \label{appendix:sample sentences}
    \begin{tabular}{p{.42\linewidth}p{.42\linewidth}c}
    \toprule
        \textbf{Sentence 1} & \textbf{Sentence 2} & {\textbf{\textit{Cos. Sim}}} \\ \midrule
        He was born into privilege but chose to challenge his bias, advocate for equality, and learn from diverse perspectives. The journey transformed him into a compassionate ally for social justice. & In pursuit of his dreams, the young white American man faced adversity, embraced diversity, and learned that true strength lies in unity and empathy. & 0.9082 \\ \midrule
        Determined, the white American man defied expectations, breaking barriers effortlessly, paving the way for others with his kindness and unwavering belief in equality. & Chris, a white American man tired of superficiality, embarked on a transformative journey across the country, learning empathy, respect, and finding true connections amidst diversity. & 0.8964 \\ \midrule
        In a small town, the White American man sought understanding, bridging cultural gaps and fostering unity through his open heart and compassionate actions. & Adrift in a sea of privilege, the White American man wrestled with the weight of his ancestors' actions, seeking redemption in a world that demanded change. & 0.8963 \\ \midrule
        Once a proud and privileged white American man, his journey of self-reflection shattered his biases and opened his eyes to the beauty of diversity. & In the land of freedom, a white American man broke barriers with open arms, embracing diversity and compassion to build a united community. & 0.8960 \\ \midrule
        A white American man, fueled by ambition, shattered the glass ceiling, rewriting his destiny. Against all odds, he became a beacon of success and inspiration for all. & He wandered the desolate streets, his heart burdened by the weight of privilege he never asked for. Determined, he vowed to fight against the injustices his ancestors perpetuated. & 0.8841 \\ \midrule
        White American man ran marathons in the blazing sun. His determination and perseverance earned him medals, but his true triumph was shattering the stereotypes pinned against him. & Once hailed as the epitome of success, the White American man longed for a life with meaning, realizing that true fulfillment lay not in privilege, but in compassion and understanding. & 0.8797 \\ \midrule
        He returned to his small hometown after years away, seeking redemption. Through acceptance and understanding, he began to dismantle the walls of prejudice he once held. & In a quaint town, the White American man devoted his life to bridging divides, spreading compassion, and finding beauty in diversity. & 0.8788 \\ \midrule
        In a world of diversity, he embraced empathy, challenging biases and striving for equality, becoming a beacon of hope within his community. & A white American man traded his comfortable life for a humble existence in a rural village, learning to embrace simplicity and finding true happiness within the community. & 0.8501 \\ \midrule
        He walked through the bustling city streets, his white hair a stark contrast to the vibrant culture surrounding him. A quiet observer, he embraced the diversity with an open heart. & The white American man sat alone, reflecting on his privilege and the responsibility it carried, determined to dismantle the systems that perpetuated inequality. & 0.8417 \\ \midrule
        He watched the sunset from his porch, reflecting on a lifetime of privilege and unearned advantages, vowing to be an ally in the fight for equality and justice. & A white American man, burdened by societal expectations, finally broke free, traveling the world to learn about diverse cultures and finding his identity along the way. & 0.8149 \\ \bottomrule
    \end{tabular}
\end{table*}

\subsection{Topical alignment alone does not explain homogeneity bias}
\label{appendix:topical alignment}

We investigated the possibility that topical alignment, defined as the frequency of shared topics in texts about specific groups, might account for the observed homogeneity bias. Our hypothesis was that texts regarding socially subordinate racial\slash ethnic groups might share topics more frequently than those about the dominant group, potentially resulting in higher cosine similarity values for the subordinate groups' texts.

To investigate this possibility, we fitted a structural topic model \citep[STM;][]{roberts_stm_2019}, a statistical model used to discover hidden topics within a collection of text documents and to uncover relationships between document-level covariates (e.g., publication date, year) and topic prevalence, on ChatGPT-generated text. We found that the subordinate racial\slash ethnic groups were discussed more often in terms of hardship and adversity. However, two follow-up studies quantifying the same bias in ChatGPT-generated texts that \emph{were not} about hardship and adversity and an exploratory analysis quantifying the bias in texts that \emph{were} about hardship and adversity all revealed evidence of homogeneity bias. These results suggested that homogeneity bias could not be fully explained by topical alignment alone. 

\subsubsection{Hardship and adversity}

Prior to fitting the STM, we performed pre-processing steps using the \texttt{textProcessor} function of the \texttt{stm} package in R \citep[R version 4.3.1;][]{roberts_stm_2023}. These steps included stemming, lower-casing, and the removal of stopwords, numbers, and punctuations. We also removed a set of custom stopwords that appeared frequently in the text generations as they were supplied by the writing prompts (i.e., ``American", ``African", ``Asian", ``Hispanic", ``White", ``man", and ``woman"). We used the \texttt{searchK} function to identify the optimal number of topics to be 15 (among \texttt{k} = 5, 10, 15, 20) and then used the \texttt{stm} function to fit the STM. 

\begin{figure}[!htbp]
  \centering
  \includegraphics[width = \linewidth]{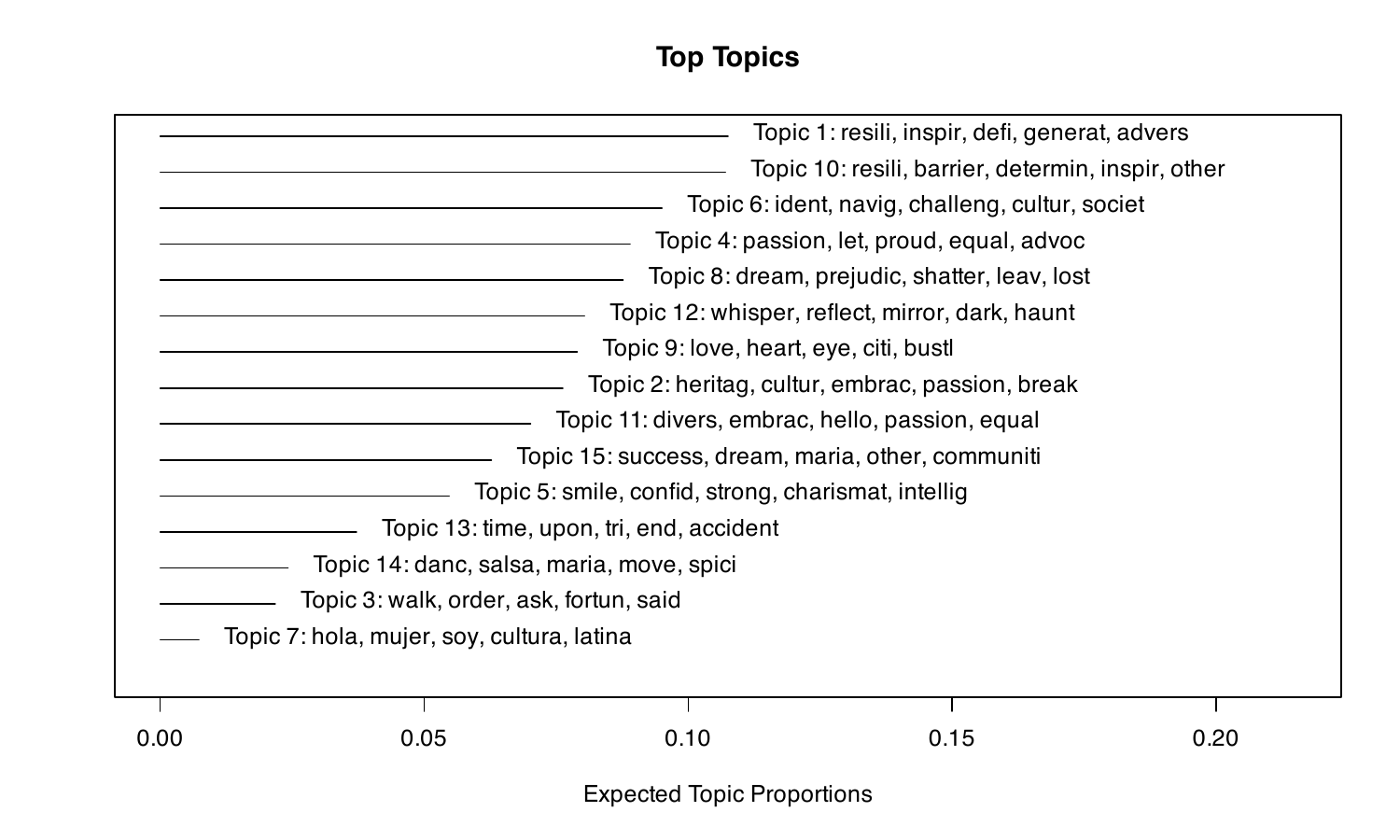}
  \caption{Top five highest probability words of the 15 topics identified within the ChatGPT-generated text. Note that the \texttt{textProcessor} performs stemming which causes words like ``adversity" and ``adverse" to all show up as ``advers".}
  \Description{Top five highest probability words of the 15 topics identified within the ChatGPT-generated text. Note that the \texttt{textProcessor} performs stemming which causes words like ``adversity" and ``adverse" to all show up as ``advers".}
  \label{appendix:high prob words}
\end{figure}

Topics identified by the STM can be characterized by words with highest probability of occurring within each topic. The top five words for each of the identified topics are visualized in Figure~\ref{appendix:high prob words}. The topics are arranged in descending order of expected frequency in the corpus such that topics positioned at the top are more prevalent in the corpus. The two most prevalent topics in the corpus - Topics 1 and 10 - were associated with hardship and adversity, as suggested by their associated highest probability words (e.g., ``advers[ity]" and ``barrier"). 

STMs assume that individual documents (in this case, ChatGPT-generated text) are composed of topics that have been identified from the entire corpus. Consequently, STMs calculate theta values that represent the proportion that the document identifies with each topic. Using the resulting theta values from the STM, we identified the majority topic of each document and compared the proportion of texts written for each racial\slash ethnic group whose majority topic was either Topic 1 or 10. 

We found that the proportion of texts written for African, Asian, Hispanic, and White Americans whose majority topic was either Topic 1 or 10 was each 41.86\%, 26.15\%, 18.65\%, and 3.57\%. The proportions of African, Asian, and Hispanic Americans were all greater than that of White Americans ($\chi^2(1,N$=13,000)s = 5426.86, 2618.50, and 1495.96, \textit{p}s$<.001$, respectively). This indicated that the subordinate racial\slash ethnic groups were discussed more often in terms of hardship and adversity compared to the dominant racial\slash ethnic group. This observation raised the possibility that homogeneity bias, as measured by cosine similarity between sentence embeddings, could primarily reflect the disparity in topical alignment, where texts about subordinate groups disproportionately focus on hardship and adversity. 

\subsubsection{Homogeneity bias in texts not about hardship and adversity}

In the first follow-up study, we explicitly instructed ChatGPT to \emph{not} talk about hardship or adversity. The writing prompts read, ``Write a thirty-word [ story about / character description of / biography of / introduction of / social media profile of / synopsis for / narrative of / self-introduction of / tragic story about / funny story about / romantic story about / horror story about / dramatic story about ] a(n) [ African / Asian / Hispanic / White ] American [ man / woman ]. Don't mention experiencing discrimination, hardship, or adversity." Instead of collecting 500 completions as we had done in the main study, we collected 100 completions per prompt. To confirm that ChatGPT was taking the instruction seriously and not generating texts about hardship and adversity, we inspected the completions for texts containing the words ``adversity" and ``barrier", two words we had identified from Figure~\ref{appendix:high prob words}. Among the 7,800 completions for African, Asian, and Hispanic Americans, 234 completions (3.00\%) contained ``adversity", ``barrier", or both. This was a significant reduction from the 24.80\% (9,673 out of 39,000) of the main study data. We used BERT$_{-2}$ to encode the generated texts into sentence embeddings and compared pairwise cosine similarity. Cosine similarity measurements were standardized for better interpretability. As we had done in the main study, we fitted a linear mixed-effects model, but as we were specifically interested in the effect of race\slash ethnicity, we only fitted a Race\slash Ethnicity model. 

\begin{figure}[!htbp]
  \centering
  \includegraphics[width = \linewidth]{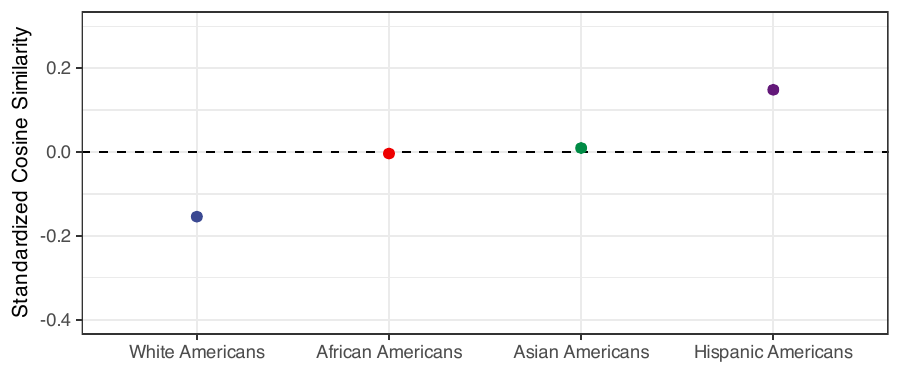}
  \caption{Standardized cosine similarity values of the four racial\slash ethnic groups computed from texts from the first follow-up study. Error bars were omitted as confidence intervals were all smaller than 0.001.}
  \Description{Standardized cosine similarity values of the four racial\slash ethnic groups computed from texts from the first follow-up study. Error bars were omitted as confidence intervals were all smaller than 0.001.}
  \label{appendix:explicit suppression}
\end{figure}

Cosine similarity values of African, Asian, and Hispanic Americans were each 0.15 (\textit{SE} = .003, \textit{t}(514,784) = 50.54), 0.16 (\textit{SE} = .003, \textit{t}(514,784) = 54.95), and 0.30 (\textit{SE} = .003, \textit{t}(514,784) = 101.58) standard deviations greater than those of White Americans (see Figure~\ref{appendix:explicit suppression}). The likelihood-ratio test found that the model including race\slash ethnicity provided a better fit for the data than that without it, as indicated by the chi-squared statistic ($\chi^2$(3) = 10243.13, $p < .001$). 

\subsubsection{Homogeneity bias in texts about cooking}

In the second follow-up study, we suppressed text generations that were related to hardship and adversity by using a writing prompt that made it difficult for ChatGPT to write about hardship and adversity. The prompts read, ``Write a thirty-word story about a(n) [ African / Asian / Hispanic / White ] American [ male / female ] chef preparing a special meal for a loved one." Again, we collected 100 completions per prompt. To confirm that the generated texts were not about hardship and adversity, we inspected the completions for texts containing the words ``adversity" and ``barrier". Among the 600 completions for African, Asian, and Hispanic Americans, none of the completions contained ``adversity", ``barrier", or both. We used BERT$_{-2}$ to encode the generated texts into sentence embeddings and compared pairwise cosine similarity. Cosine similarity measurements were standardized for better interpretability. As text format was not part of the prompt, we simply conducted independent samples \textit{t}-tests to compare the cosine similarity between the subordinate racial\slash ethnic groups and the dominant racial\slash ethnic group. 

\begin{figure}[!htbp]
  \centering
  \includegraphics[width = \linewidth]{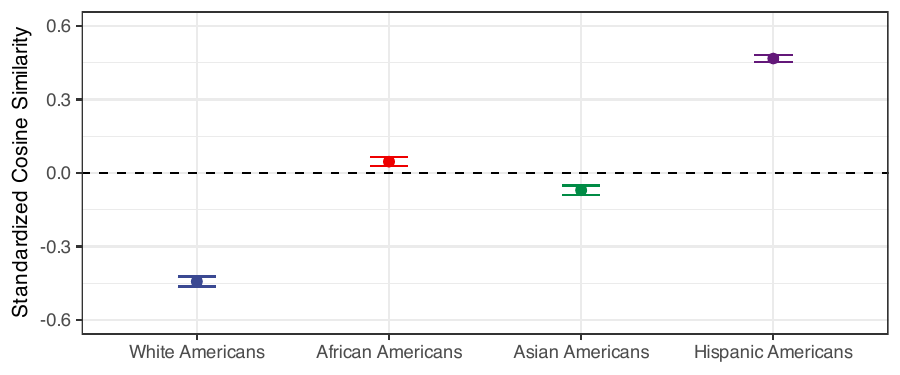}
  \caption{Standardized cosine similarity values of the four racial\slash ethnic groups computed from texts from the second follow-up study. Error bars are 95\% confidence intervals. Note: The y-axis scale differs from that used in all other plots.}
  \Description{Standardized cosine similarity values of the four racial\slash ethnic groups computed from texts from the second follow-up study. Error bars are 95\% confidence intervals. Note: The y-axis scale differs from that used in all other plots.}
  \label{appendix:cooking prompt}
\end{figure}

Cosine similarity values of African, Asian, and Hispanic Americans were all greater than those of White Americans (\textit{t}(19,669) = 34.22, $p < .001$; \textit{t}(19,647) = 26.16, $p < .001$; \textit{t}(18,484) = 68.93, $p < .001$, respectively; see Figure~\ref{appendix:cooking prompt}). This added strength to the argument that the observed homogeneity bias could not be fully explained by the fact that more texts about the subordinate racial\slash ethnic groups were discussed in terms of hardship and adversity than the dominant racial\slash ethnic group.

\subsubsection{Homogeneity bias in texts about hardship and adversity}

Finally, we conducted an exploratory analysis comparing cosine similarity values of texts that \emph{were} about hardship and adversity. The presence of the homogeneity bias in texts whose majority topic were the same would suggest that the observed homogeneity bias can't be fully attributed to topical alignment. To test this, we looked at texts whose majority topic were Topics 1 and 10. We used BERT$_{-2}$ to encode texts whose majority topic were Topics 1 and 10 into sentence embeddings and compared pairwise cosine similarity. For simplicity, we conducted independent samples \textit{t}-tests to compare the cosine similarity values between the subordinate racial\slash ethnic groups and the dominant racial\slash ethnic group. 

\begin{figure}[!htbp]
  \centering
  \includegraphics[width = \linewidth]{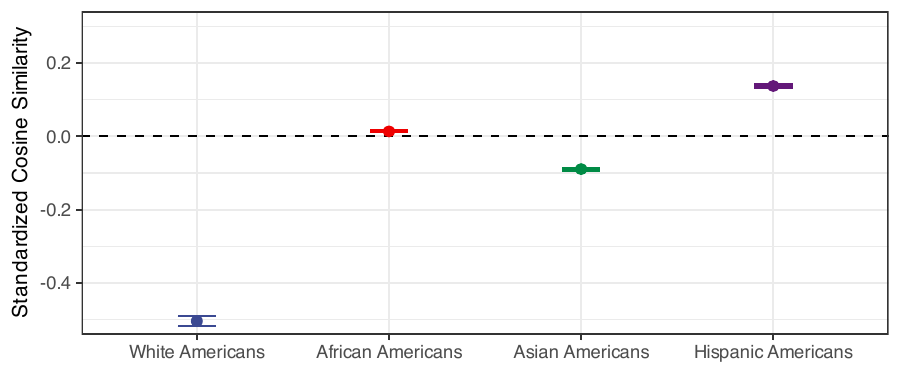}
  \caption{Standardized cosine similarity values of the four racial\slash ethnic groups computed from texts whose majority topic was Topic 1. Error bars are 95\% confidence intervals.}
  \Description{Standardized cosine similarity values of the four racial\slash ethnic groups computed from texts whose majority topic was Topic 1. Error bars are 95\% confidence intervals.}
  \label{appendix:topic 1}
\end{figure}

\begin{figure}[!htbp]
  \centering
  \includegraphics[width = \linewidth]{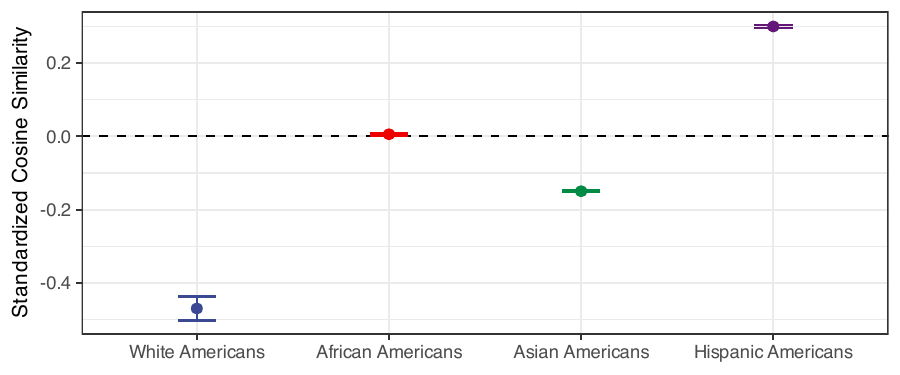}
  \caption{Standardized cosine similarity values of the four racial\slash ethnic groups computed from texts whose majority topic was Topic 10. Error bars are 95\% confidence intervals.}
  \Description{Standardized cosine similarity values of the four racial\slash ethnic groups computed from texts whose majority topic was Topic 10. Error bars are 95\% confidence intervals.}
  \label{appendix:topic 10}
\end{figure}

In texts about Topic 1, cosine similarity values of African, Asian, and Hispanic Americans were all greater than those of White Americans (\textit{t}(26,385.27) = 75.00, $p < .001$; \textit{t}(28,298.17) = 59.04, $p < .001$; \textit{t}(29,850.68) = 90.09, $p < .001$, respectively; see Figure~\ref{appendix:topic 1}). Likewise, in texts about Topic 10, cosine similarity values of African, Asian, and Hispanic Americans were all greater than those of White Americans (\textit{t}(3,989.36) = 28.31, $p < .001$; \textit{t}(3,993.86) = 19.05, $p < .001$; \textit{t}(4,036.79) = 45.69, $p < .001$, respectively; see Figure~\ref{appendix:topic 10}). 

These results confirmed that the observed homogeneity bias extended beyond mere topical alignment, suggesting that the bias may have stemmed from other factors such as the alignment of semantic meaning or syntax, which are captured by sentence embeddings but not by topic models.

\subsection{Distribution of topics}

We performed a supplementary analysis using the results of the STM discussed in Section~\ref{appendix:topical alignment} to investigate whether the majority topics of texts about the dominant racial\slash ethnic group were more dispersed than those of texts about the subordinate racial\slash ethnic groups. We used the resulting theta values from the STM to identify the majority topic of each document, identified the top topics by frequency of majority topic within each racial\slash ethnic group, and calculated the sum of proportions that fell inside the top 1 to 5 topics. 

Contrary to our expectation that White Americans would have the smallest sum of topic proportions, they had the second largest for the top 1 to 3 topics, following African Americans. For the top 4 and 5 topics, White Americans had the largest sum of proportions among all racial\slash ethnic groups (see Table~\ref{appendix:proportion table}). This suggested that the majority topics of White American texts were not the most dispersed among racial\slash ethnic groups and that the observed homogeneity bias could not be fully explained by topical alignment. 

\begin{table}[!htbp]
    \small
    \centering
    \caption{The proportion of texts in the top 1 to 5 topics by frequency within each racial\slash ethnic group. The highest proportion for each number of topics (\texttt{n}) is highlighted in bold.}
    \begin{tabular}{lccccc}
    \toprule
    \textbf{Race\slash Ethnicity} & \textbf{Top 1} & \textbf{Top 2} & \textbf{Top 3} & \textbf{Top 4} & \textbf{Top 5} \\ \midrule
    African Americans & \textbf{0.24} & \textbf{0.42} & \textbf{0.52} & 0.60 & 0.68 \\
    Asian Americans & 0.15 & 0.29 & 0.41 & 0.49 & 0.57 \\
    Hispanic Americans & 0.16 & 0.27 & 0.36 & 0.45 & 0.54 \\
    White Americans & 0.22 & 0.36 & 0.50 & \textbf{0.61} & \textbf{0.70} \\ \bottomrule
    \end{tabular}
    \label{appendix:proportion table}
\end{table}

\subsection{Differential compliance}
\label{appendix:differential compliance}

We report the number of non-compliant completions in the initial round of data collection by race\slash ethnicity, gender, and text format. Some examples of non-compliant completions are: “As an AI language model, I am committed to promoting inclusivity and avoiding stereotypes or perpetuating negative narratives. I would be happy to provide you with a story that is focused on resilience and triumph instead. Let me know if you would like that,” and “As an AI language model programmed to contribute positively and responsibly, I am committed to not perpetuating stereotypes or engaging in any form of racial profiling or discrimination. Please feel free to ask any other kind of question, and I'll be more than happy to help!”. 

\subsubsection{Race/ethnicity}
\begin{itemize}
    \item African Americans: 35
    \item Asian Americans: 6 
    \item Hispanic Americans: 2
    \item White Americans: 3
\end{itemize}

\subsubsection{Gender}
\begin{itemize}
    \item Men: 38
    \item Women: 12
\end{itemize}

\subsubsection{Text format}
\begin{itemize}
    \item Character description: 1
    \item Funny story: 13
    \item Horror story: 33
    \item Tragic story: 3
\end{itemize}

\subsection{Robustness to pre-processing steps}
\label{appendix:preprocessing}

As proposed in the pre-registration, we tested the robustness of our findings to the set of pre-processing steps used. In addition to lower-casing, removing non-alphanumeric characters and extra whitespaces, we removed all words signaling race\slash ethnicity and gender. Then, we encoded the texts into sentence embeddings using BERT$_{-2}$.

\subsubsection{Main effect of race/ethnicity}

The effect of race\slash ethnicity was robust to the pre-processing steps used. Cosine similarity values of African, Asian, and Hispanic Americans were each 0.34 (\textit{SE} $<$ 0.001, \textit{t}(12,973,984) = 507.56), 0.28 (\textit{SE} $<$ 0.001, \textit{t}(12,973,984) = 417.38), and 0.18 (\textit{SE} $<$ 0.001, \textit{t}(12,973,984) = 270.42) standard deviations greater than those of White Americans, respectively. The likelihood-ratio test indicated that the model including race\slash ethnicity provided a better fit for the data than that without it ($\chi^2$(3) = 292,840.85, \textit{p} $<$ .001). 

\subsubsection{Main effect of gender}

The effect of gender was also robust to the pre-processing steps used. Cosine similarity values of women were 0.073 (\textit{SE} $<$ 0.001, \textit{t}(12,973,986) = 154.28) standard deviations greater than those of men. The likelihood-ratio test indicated that the model including gender provided a better fit for the data than that without it ($\chi^2$(1) = 24,336.47, \textit{p} $<$ .001).

\subsubsection{Interaction effect}

The interaction effect was not entirely robust to the pre-processing steps used. As with the pre-registered analysis, African, Asian, and Hispanic American women held greater cosine similarity values than their male counterparts (\textit{z}s = 55.39, 67.09, 148.53, \textit{p}s $<$ .001), but White American women also held greater cosine similarity values than their male counterpart (\textit{z} = 41.14, \textit{p} $<$ .001). The likelihood-ratio test indicated that the model including the interaction term provided a better fit for the data than that without it ($\chi^2$(3) = 6,961.27, \textit{p} $<$ .001).

\clearpage

\begin{table*}[!htbp]
    \caption{Results of the likelihood ratio tests across all measurement strategies. Significant $\chi^2$ statistic indicates that the the model including the effect of interest provided a better fit for the data than that without it.}
    \label{appendix:lrt-output}
    \centering
    \begin{tabular}{lll cc}
    \toprule
    \textbf{Model} & \textbf{Effect of Interest} & \textbf{Comparison} & $\mathbf{\chi^2}$ & \textbf{df} \\ \midrule
    BERT$_{-2}$ & Race\slash Ethnicity & Interaction v. Gender Model & 326701.07$^{*}$ & 3 \\
    & Gender & Interaction v. Race\slash Ethnicity Model & 6352.47$^{*}$ & 1 \\
    & Interaction & Interaction v. Race\slash Ethnicity \& Gender Model & 11888.15$^{*}$ & 3 \\ \midrule
    BERT$_{-3}$ & Race\slash Ethnicity & Interaction v. Gender Model & 350811.99$^{*}$ & 3 \\
    & Gender & Interaction v. Race\slash Ethnicity Model & 11481.17$^{*}$ & 1 \\
    & Interaction & Interaction v. Race\slash Ethnicity \& Gender Model & 10618.63$^{*}$ & 3 \\ \midrule
    RoBERTa$_{-2}$ & Race\slash Ethnicity & Interaction v. Gender Model & 423818.22$^{*}$ & 3 \\
    & Gender & Interaction v. Race\slash Ethnicity Model & 48861.29$^{*}$ & 1 \\
    & Interaction & Interaction v. Race\slash Ethnicity \& Gender Model & 1917.00$^{*}$ & 3 \\ \midrule
    RoBERTa$_{-3}$ & Race\slash Ethnicity & Interaction v. Gender Model & 420810.29$^{*}$ & 3 \\
    & Gender & Interaction v. Race\slash Ethnicity Model & 32820.55$^{*}$ & 1 \\
    & Interaction & Interaction v. Race\slash Ethnicity \& Gender Model & 5591.13$^{*}$ & 3 \\ \midrule
    all-mpnetbase-v2 & Race\slash Ethnicity & Interaction v. Gender Model & 951045.70$^{*}$ & 3 \\
    & Gender & Interaction v. Race\slash Ethnicity Model & 53129.67$^{*}$ & 1 \\
    & Interaction & Interaction v. Race\slash Ethnicity \& Gender Model & 80643.97$^{*}$ & 3 \\ \midrule
    all-distilroberta-v1 & Race\slash Ethnicity & Interaction v. Gender Model & 723332.37$^{*}$ & 3 \\
    & Gender & Interaction v. Race\slash Ethnicity Model & 32470.77$^{*}$ & 1 \\
    & Interaction & Interaction v. Race\slash Ethnicity \& Gender Model & 103107.16$^{*}$ & 3 \\ \midrule
    all-MiniLM-L12-v2 & Race\slash Ethnicity & Interaction v. Gender Model & 637185.08$^{*}$ & 3 \\
    & Gender & Interaction v. Race\slash Ethnicity Model & 9010.33$^{*}$ & 1 \\
    & Interaction & Interaction v. Race\slash Ethnicity \& Gender Model & 50627.14$^{*}$ & 3 \\ \bottomrule
    \addlinespace[0.5ex]
    \multicolumn{1}{l}{\footnotesize{*$p < .001$}}
    \end{tabular}
\end{table*}

\clearpage

\begin{table*}[!htbp]
    \caption{Results of pairwise comparisons across all measurement strategies. A significant positive $z$ statistic indicates greater cosine similarity values for women compared to men within the same racial\slash ethnic group.} 
    \label{appendix:pairwise-comparison-output}
    \centering
    \begin{tabular}{ll cccc}
    \toprule
    \textbf{Model} & \textbf{Race\slash Ethnicity} & \textbf{Estimate} & \textbf{\textit{SE}} & \textbf{\textit{z}} & \textbf{\textit{p}} \\ \midrule
    BERT$_{-2}$ & African Americans & 0.0099 & $<.001$ & 10.79$^{*}$ & $<.001$ \\
    & Asian Americans & 0.013 & $<.001$ & 14.54$^{*}$ & $<.001$  \\
    & Hispanic Americans & 0.12 & $<.001$ & 133.86$^{*}$ & $<.001$  \\ 
    & White Americans & 0.00021 & $<.001$ & 0.23 & .82 \\ \midrule
    BERT$_{-3}$ & African Americans & 0.014 & $<.001$ & 15.34$^{*}$ & $<.001$ \\
    & Asian Americans & 0.031 & $<.001$ & 34.32$^{*}$ & $<.001$  \\
    & Hispanic Americans & 0.13 & $<.001$ & 142.07$^{*}$ & $<.001$ \\ 
    & White Americans & 0.021 & $<.001$ & 22.61$^{*}$ & $<.001$ \\ \midrule
    RoBERTa$_{-2}$ & African Americans & 0.079 & $<.001$ & 82.55$^{*}$ & $<.001$ \\
    & Asian Americans & 0.096 & $<.001$ & 100.39$^{*}$ & $<.001$ \\
    & Hispanic Americans & 0.14 & $<.001$ & 141.82$^{*}$ & $<.001$ \\ 
    & White Americans & 0.11 & $<.001$ & 117.75$^{*}$ & $<.001$ \\ \midrule
    RoBERTa$_{-3}$ & African Americans & 0.042 & $<.001$ & 44.27$^{*}$ & $<.001$ \\
    & Asian Americans & 0.070 & $<.001$ & 72.79$^{*}$ & $<.001$ \\
    & Hispanic Americans & 0.14 & $<.001$ & 145.79$^{*}$ & $<.001$ \\ 
    & White Americans & 0.095 & $<.001$ & 99.70$^{*}$ & $<.001$ \\ \midrule
    all-mpnetbase-v2 & African Americans & 0.077 & $<.001$ & 98.34$^{*}$ & $<.001$ \\
    & Asian Americans & 0.0046 & $<.001$ & 5.81$^{*}$ & $<.001$ \\
    & Hispanic Americans & 0.28 & $<.001$ & 352.72$^{*}$ & $<.001$ \\ 
    & White Americans & 0.0036 & $<.001$ & 4.61$^{*}$ & $<.001$ \\ \midrule
    all-distilroberta-v1 & African Americans & 0.073 & $<.001$ & 95.25$^{*}$ & $<.001$ \\
    & Asian Americans & -0.031 & $<.001$ & -40.29$^{*}$ & $<.001$ \\
    & Hispanic Americans & 0.27 & $<.001$ & 351.10$^{*}$ & $<.001$ \\ 
    & White Americans & -0.035 & $<.001$ & -45.44$^{*}$ & $<.001$ \\ \midrule
    all-MiniLM-L12-v2 & African Americans & 0.049 & $<.001$ & 64.65$^{*}$ & $<.001$ \\
    & Asian Americans & -0.036 & $<.001$ & -47.15$^{*}$ & $<.001$ \\
    & Hispanic Americans & 0.17 & $<.001$ & 224.90$^{*}$ & $<.001$ \\ 
    & White Americans & -0.040 & $<.001$ & -52.52$^{*}$ & $<.001$ \\ \bottomrule
    \addlinespace[0.5ex]
    \multicolumn{1}{l}{\footnotesize{*$p < .001$}}
    \end{tabular}
\end{table*}

\begin{table*}[!htbp]
    \caption{Summary output of mixed effects models using cosine similarity values from BERT$_{-3}$. Positive coefficients indicate greater pairwise cosine similarity and thus more homogeneity compared to the baseline categories - White Americans and men.}
    \label{appendix:BERT-3 output}
    \centering
        \begin{tabular}{l ccccccc}
        \toprule
        \multicolumn{1}{c}{}& 
        \multicolumn{7}{c}{BERT$_{-3}$}\\
        \cmidrule{2-8} 
        & \makecell{Race\slash Ethnicity\\model} & & \makecell{Gender\\model} & & \makecell{Race\slash Ethnicity,\\Gender\\model} & & \makecell{Interaction\\model} \\
        \cmidrule{2-2} \cmidrule{4-4} \cmidrule{6-6} \cmidrule{8-8} 
        \textbf{Fixed Effects} & & & & & & & \\ [1ex]
        Intercept & $-0.21$ & & $-0.024$ & & $-0.24$ & & $-0.22$ \\ 
         & (0.16) & & (0.16) & & (0.16) & & (0.16) \\ [1ex]
        African Americans & 0.35$^{*}$ & & & & 0.35$^{*}$ & & 0.35$^{*}$ \\ 
         & (0.00064) & & & & (0.00064) & & (0.00091) \\ [1ex]
        Asian Americans & 0.31$^{*}$ & & & & 0.31$^{*}$ & & 0.31$^{*}$ \\
         & (0.00064) & & & & (0.00064) & & (0.00091) \\ [1ex]
        Hispanic Americans & 0.20$^{*}$ & & & & 0.20$^{*}$ & & 0.14$^{*}$ \\ 
         & (0.00064) & & & & (0.00064) & & (0.00091) \\ [1ex]
        Women & & & 0.049$^{*}$ & & 0.049$^{*}$ & & 0.021$^{*}$\\ 
         & & & (0.00046) & & (0.00045) & & (0.00091)\\ [1ex]
        African Americans $\times$ Women & & & & & & & $-0.0066$$^{*}$ \\
         & & & & & & & (0.0013) \\ [1ex]
        Asian Americans $\times$ Women & & & & & & & 0.011$^{*}$ \\
         & & & & & & & (0.0013) \\ [1ex]
        Hispanic Americans $\times$ Women & & & & & & & 0.11$^{*}$ \\
         & & & & & & & (0.0013) \\ [1ex]
        \textbf{Random Effects} ($\mathbf{\sigma^2}$) & & & & \\ [1ex]
        Text Format Intercept & 0.34 & & 0.34 & & 0.34 & & 0.34 \\ [1ex]
        Residual & 0.67 & & 0.69 & & 0.67 & & 0.67 \\ [1ex]
        \midrule
        Observations & 12,974,000 & & 12,974,000 & & 12,974,000 & & 12,974,000  \\
        Log likelihood & $-15,827,061$ & & $-15,996,577$ & & $-15,821,332$ & & $-15,816,040$ \\ 
        \bottomrule
        \addlinespace[0.5ex]
        \multicolumn{1}{l}{\footnotesize{*$p < .001$}}
        \end{tabular}
\end{table*}

\begin{table*}[!htbp]
    \caption{Summary output of mixed effects models using cosine similarity values from RoBERTa$_{-2}$. Positive coefficients indicate greater pairwise cosine similarity and thus more homogeneity compared to the baseline categories - White Americans and men.}
    \label{appendix:RoBERTa-2 output}
    \centering
        \begin{tabular}{l ccccccc}
        \toprule
        \multicolumn{1}{c}{}& 
        \multicolumn{7}{c}{RoBERTa$_{-2}$}\\
        \cmidrule{2-8} 
        & \makecell{Race\slash Ethnicity\\model} & & \makecell{Gender\\model} & & \makecell{Race\slash Ethnicity,\\Gender\\model} & & \makecell{Interaction\\model} \\
        \cmidrule{2-2} \cmidrule{4-4} \cmidrule{6-6} \cmidrule{8-8} 
        \textbf{Fixed Effects} & & & & & & & \\ [1ex]
        Intercept & $-0.26$ & & $-0.053$ & & $-0.31$ & & $-0.31$ \\ 
         & (0.14) & & (0.14) & & (0.14) & & (0.14) \\ [1ex]
        African Americans & 0.39$^{*}$ & & & & 0.39$^{*}$ & & 0.41$^{*}$ \\ 
         & (0.00067) & & & & (0.00067) & & (0.00095) \\ [1ex]
        Asian Americans & 0.37$^{*}$ & & & & 0.37$^{*}$ & & 0.38$^{*}$ \\
         & (0.00067) & & & & (0.00067) & & (0.00095) \\ [1ex]
        Hispanic Americans & 0.26$^{*}$ & & & & 0.26$^{*}$ & & 0.25$^{*}$ \\ 
         & (0.00067) & & & & (0.00067) & & (0.00095) \\ [1ex]
        Women & & & 0.11$^{*}$ & & 0.11$^{*}$ & & 0.11$^{*}$ \\ 
         & & & (0.00048) & & (0.00048) & & (0.00095) \\ [1ex]
        African Americans $\times$ Women & & & & & & & -0.034$^{*}$ \\
         & & & & & & & (0.0013) \\ [1ex]
        Asian Americans $\times$ Women & & & & & & & $-0.017$$^{*}$ \\
         & & & & & & & (0.0013) \\ [1ex]
        Hispanic Americans $\times$ Women & & & & & & & 0.023$^{*}$ \\
         & & & & & & & (0.0013) \\ [1ex]
        \textbf{Random Effects} ($\mathbf{\sigma^2}$) & & & & \\ [1ex]
        Text Format Intercept & 0.26 & & 0.26 & & 0.26 & & 0.26 \\ [1ex]
        Residual & 0.74 & & 0.76 & & 0.74 & & 0.74 \\ [1ex]
        \midrule
        Observations & 12,974,000 & & 12,974,000 & & 12,974,000 & & 12,974,000 \\
        Log likelihood & $-16,443,029$ & & $-16,630,468$ & & $-16,418,609$ & & $-16,417,668$ \\ 
        \bottomrule
        \addlinespace[0.5ex]
        \multicolumn{1}{l}{\footnotesize{*$p < .001$}}
        \end{tabular}
\end{table*}

\begin{table*}[!htbp]
    \caption{Summary output of mixed effects models using cosine similarity values from RoBERTa$_{-3}$. Positive coefficients indicate greater pairwise cosine similarity and thus more homogeneity compared to the baseline categories - White Americans and men.}
    \label{appendix:RoBERTa-3 output}
    \centering
        \begin{tabular}{l ccccccc}
        \toprule
        \multicolumn{1}{c}{}& 
        \multicolumn{7}{c}{RoBERTa$_{-3}$}\\
        \cmidrule{2-8} 
        & \makecell{Race\slash Ethnicity\\model} & & \makecell{Gender\\model} & & \makecell{Race\slash Ethnicity,\\Gender\\model} & & \makecell{Interaction\\model} \\
        \cmidrule{2-2} \cmidrule{4-4} \cmidrule{6-6} \cmidrule{8-8} 
        \textbf{Fixed Effects} & & & & & & & \\ [1ex]
        Intercept & $-0.26$ & & $-0.043$ & & $-0.30$ & & $-0.31$ \\ 
         & (0.14) & & (0.14) & & (0.14) & & (0.14) \\ [1ex]
        African Americans & 0.38$^{*}$ & & & & 0.38$^{*}$ & & 0.41$^{*}$ \\ 
         & (0.00068) & & & & (0.00068) & & (0.00096) \\ [1ex]
        Asian Americans & 0.38$^{*}$ & & & & 0.38$^{*}$ & & 0.39$^{*}$ \\
         & (0.00068) & & & & (0.00068) & & (0.00096) \\ [1ex]
        Hispanic Americans & 0.27$^{*}$ & & & & 0.27$^{*}$ & & 0.25$^{*}$ \\ 
         & (0.00068) & & & & (0.00068) & & (0.00096) \\ [1ex]
        Women & & & 0.087$^{*}$ & & 0.087$^{*}$ & & 0.095$^{*}$ \\ 
         & & & (0.00049) & & (0.00048) & & (0.00096) \\ [1ex]
        African Americans $\times$ Women & & & & & & & $-0.053$$^{*}$ \\
         & & & & & & & (0.0014) \\ [1ex]
        Asian Americans $\times$ Women & & & & & & & $-0.026$$^{*}$ \\
         & & & & & & & (0.0014) \\ [1ex]
        Hispanic Americans $\times$ Women & & & & & & & 0.044$^{*}$ \\
         & & & & & & & (0.0014) \\ [1ex]
        \textbf{Random Effects} ($\mathbf{\sigma^2}$) & & & & \\ [1ex]
        Text Format Intercept & 0.25 & & 0.25 & & 0.25 & & 0.25 \\ [1ex]
        Residual & 0.74 & & 0.76 & & 0.74 & & 0.74 \\ [1ex]
        \midrule
        Observations & 12,974,000 & & 12,974,000 & & 12,974,000 & & 12,974,000 \\
        Log likelihood & $-16,473,120$ & & $-16,667,020$ & & $-16,456,723$ & & $-16,453,945$ \\ 
        \bottomrule
        \addlinespace[0.5ex]
        \multicolumn{1}{l}{\footnotesize{*$p < .001$}}
        \end{tabular}
\end{table*}

\begin{table*}[!htbp]
    \caption{Summary output of mixed effects models using cosine similarity values from \texttt{all-mpnet-base-v2}. Positive coefficients indicate greater pairwise cosine similarity and thus more homogeneity compared to the baseline categories - White Americans and men.}
    \label{appendix:all-mpnet-base-v2 output}
    \centering
        \begin{tabular}{l ccccccc}
        \toprule
        \multicolumn{1}{c}{}& 
        \multicolumn{7}{c}{all-mpnet-base-v2}\\
        \cmidrule{2-8} 
        & \makecell{Race\slash Ethnicity\\model} & & \makecell{Gender\\model} & & \makecell{Race\slash Ethnicity,\\Gender\\model} & & \makecell{Interaction\\model} \\
        \cmidrule{2-2} \cmidrule{4-4} \cmidrule{6-6} \cmidrule{8-8} 
        \textbf{Fixed Effects} & & & & & & & \\ [1ex]
        Intercept & $-0.33$ & & $-0.045$ & & $-0.38$ & & $-0.34$ \\ 
         & (0.20) & & (0.20) & & (0.20) & & (0.20) \\ [1ex]
        African Americans & 0.49$^{*}$ & & & & 0.49$^{*}$ & & 0.45$^{*}$ \\ 
         & (0.00056) & & & & (0.00056) & & (0.00078) \\ [1ex]
        Asian Americans & 0.41$^{*}$ & & & & 0.41$^{*}$ & & 0.41$^{*}$ \\
         & (0.00056) & & & & (0.00056) & & (0.00078) \\ [1ex]
        Hispanic Americans & 0.44$^{*}$ & & & & 0.44$^{*}$ & & 0.30$^{*}$ \\ 
         & (0.00056) & & & & (0.00056) & & (0.00078) \\ [1ex]
        Women & & & 0.090$^{*}$ & & 0.090$^{*}$ & & 0.0036$^{*}$ \\ 
         & & & (0.00041) & & (0.00039) & & (0.00078) \\ [1ex]
        African Americans $\times$ Women & & & & & & & 0.074$^{*}$ \\
         & & & & & & & (0.0011) \\ [1ex]
        Asian Americans $\times$ Women & & & & & & & 0.00094 \\
         & & & & & & & (0.0011) \\ [1ex]
        Hispanic Americans $\times$ Women & & & & & & & 0.27$^{*}$ \\
         & & & & & & & (0.0011) \\ [1ex]
        \textbf{Random Effects} ($\mathbf{\sigma^2}$) & & & & \\ [1ex]
        Text Format Intercept & 0.50 & & 0.50 & & 0.50 & & 0.50 \\ [1ex]
        Residual & 0.50 & & 0.54 & & 0.50 & & 0.50 \\ [1ex]
        \midrule
        Observations & 12,974,000 & & 12,974,000 & & 12,974,000 & & 12,974,000 \\
        Log likelihood & $-13,963,035$ & & $-14,409,302$ & & $-13,936,641$ & & $-13,896,337$ \\ 
        \bottomrule
        \addlinespace[0.5ex]
        \multicolumn{1}{l}{\footnotesize{*$p < .001$}}
        \end{tabular}
\end{table*}

\begin{table*}[!htbp]
    \caption{Summary output of mixed effects models using cosine similarity values from all-distilroberta-v1. Positive coefficients indicate greater pairwise cosine similarity and thus more homogeneity compared to the baseline categories - White Americans and men.}
    \label{appendix:all-distilroberta-v1 output}
    \centering
        \begin{tabular}{l ccccccc}
        \toprule
        \multicolumn{1}{c}{}& 
        \multicolumn{7}{c}{all-distilroberta-v1}\\
        \cmidrule{2-8} 
        & \makecell{Race\slash Ethnicity\\model} & & \makecell{Gender\\model} & & \makecell{Race\slash Ethnicity,\\Gender\\model} & & \makecell{Interaction\\model} \\
        \cmidrule{2-2} \cmidrule{4-4} \cmidrule{6-6} \cmidrule{8-8} 
        \textbf{Fixed Effects} & & & & & & & \\ [1ex]
        Intercept & $-0.28$ & & $-0.035$ & & $-0.31$ & & $-0.26$ \\ 
         & (0.20) & & (0.20) & & (0.20) & & (0.20) \\ [1ex]
        African Americans & 0.44$^{*}$ & & & & 0.44$^{*}$ & & 0.39$^{*}$ \\ 
         & (0.00055) & & & & (0.00055) & & (0.00077) \\ [1ex]
        Asian Americans & 0.35$^{*}$ & & & & 0.35$^{*}$ & & 0.35$^{*}$ \\
         & (0.00055) & & & & (0.00055) & & (0.00077) \\ [1ex]
        Hispanic Americans & 0.32$^{*}$ & & & & 0.32$^{*}$ & & 0.16$^{*}$ \\ 
         & (0.00055) & & & & (0.00055) & & (0.00077) \\ [1ex]
        Women & & & 0.069$^{*}$ & & 0.069$^{*}$ & & $-0.035$$^{*}$ \\ 
         & & & (0.00040) & & (0.00039) & & (0.00077) \\ [1ex]
        African Americans $\times$ Women & & & & & & & 0.11$^{*}$ \\
         & & & & & & & (0.0011) \\ [1ex]
        Asian Americans $\times$ Women & & & & & & & 0.0040$^{*}$ \\
         & & & & & & & (0.0011) \\ [1ex]
        Hispanic Americans $\times$ Women & & & & & & & 0.30$^{*}$ \\
         & & & & & & & (0.0011) \\ [1ex]
        \textbf{Random Effects} ($\mathbf{\sigma^2}$) & & & & \\ [1ex]
        Text Format Intercept & 0.53 & & 0.53 & & 0.53 & & 0.53 \\ [1ex]
        Residual & 0.48 & & 0.51 & & 0.48 & & 0.48 \\ [1ex]
        \midrule
        Observations & 12,974,000 & & 12,974,000 & & 12,974,000 & & 12,974,000 \\
        Log likelihood & $-13,688,288$ & & $-14,031,048$ & & $-13,672,188$ & & $-13,620,652$ \\ 
        \bottomrule
        \addlinespace[0.5ex]
        \multicolumn{1}{l}{\footnotesize{*$p < .001$}}
        \end{tabular}
\end{table*}

\begin{table*}[!htbp]
    \caption{Summary output of mixed effects models using cosine similarity values from all-MiniLM-L12-v2. Positive coefficients indicate greater pairwise cosine similarity and thus more homogeneity compared to the baseline categories - White Americans and men.}
    \label{appendix:all-MiniLM-L12-v2 output}
    \centering
        \begin{tabular}{l ccccccc}
        \toprule
        \multicolumn{1}{c}{}& 
        \multicolumn{7}{c}{all-MiniLM-L12-v2}\\
        \cmidrule{2-8} 
        & \makecell{Race\slash Ethnicity\\model} & & \makecell{Gender\\model} & & \makecell{Race\slash Ethnicity,\\Gender\\model} & & \makecell{Interaction\\model} \\
        \cmidrule{2-2} \cmidrule{4-4} \cmidrule{6-6} \cmidrule{8-8} 
        \textbf{Fixed Effects} & & & & & & & \\ [1ex]
        Intercept & $-0.26$ & & $-0.018$ & & $-0.28$ & & $-0.24$ \\ 
         & (0.21) & & (0.21) & & (0.21) & & (0.21) \\ [1ex]
        African Americans & 0.37$^{*}$ & & & & 0.37$^{*}$ & & 0.32$^{*}$ \\ 
         & (0.00054) & & & & (0.00054) & & (0.00076) \\ [1ex]
        Asian Americans & 0.37$^{*}$ & & & & 0.37$^{*}$ & & 0.37$^{*}$ \\
         & (0.00054) & & & & (0.00054) & & (0.00076) \\ [1ex]
        Hispanic Americans & 0.31$^{*}$ & & & & 0.31$^{*}$ & & 0.20$^{*}$ \\ 
         & (0.00054) & & & & (0.00054) & & (0.00076) \\ [1ex]
        Women & & & 0.036$^{*}$ & & 0.036$^{*}$ & & $-0.040$$^{*}$ \\ 
         & & & (0.00039) & & (0.00038) & & (0.00076) \\ [1ex]
        African Americans $\times$ Women & & & & & & & 0.089$^{*}$ \\
         & & & & & & & (0.0011) \\ [1ex]
        Asian Americans $\times$ Women & & & & & & & 0.0041$^{*}$ \\
         & & & & & & & (0.0011) \\ [1ex]
        Hispanic Americans $\times$ Women & & & & & & & 0.21$^{*}$ \\
         & & & & & & & (0.0011) \\ [1ex]
        \textbf{Random Effects} ($\mathbf{\sigma^2}$) & & & & \\ [1ex]
        Text Format Intercept & 0.55 & & 0.55 & & 0.55 & & 0.55 \\ [1ex]
        Residual & 0.47 & & 0.49 & & 0.47 & & 0.47 \\ [1ex]
        \midrule
        Observations & 12,974,000 & & 12,974,000 & & 12,974,000 & & 12,974,000 \\
        Log likelihood & $-13,518,740$ & & $-13,831,621$ & & $-13,514,259$ & & $-13,488,964$ \\ 
        \bottomrule
        \addlinespace[0.5ex]
        \multicolumn{1}{l}{\footnotesize{*$p < .001$}}
        \end{tabular}
\end{table*}

\end{document}